\documentclass{article}

\usepackage{PRIMEarxiv}

\usepackage[utf8]{inputenc} 
\usepackage[T1]{fontenc}    
\usepackage{hyperref}       
\usepackage{url}            
\usepackage{booktabs}       
\usepackage{amsfonts}       
\usepackage{nicefrac}       
\usepackage{microtype}      
\usepackage{lipsum}
\usepackage{fancyhdr}       
\usepackage{graphicx}       
\usepackage{amsmath,amsfonts}
\usepackage{algorithmic}
\usepackage{algorithm}
\usepackage{array}
\usepackage{cuted}
\usepackage{textcomp}
\usepackage{stfloats}
\usepackage{verbatim}
\usepackage{epsfig}
\usepackage{amssymb}
\usepackage{subcaption}
\usepackage{multirow}

\usepackage{pifont}
\usepackage{float}
\usepackage{breqn}
\usepackage[normalem]{ulem} 

\newcommand{\METHOD}{BAdd}

\usepackage{authblk}

\graphicspath{{figures/}}

\title{BAdd: Bias Mitigation through Bias Addition} 

\author[1,2]{Ioannis Sarridis} 
\author[1]{Christos Koutlis}
\author[1]{Symeon Papadopoulos}
\author[2]{Christos Diou}
\affil[1]{CERTH-ITI, Thessaloniki, Greece}
\affil[2]{Harokopio University of Athens, Athens, Greece}
\affil[ ]{\texttt{\{gsarridis,ckoutlis,papadop\}@iti.gr, cdiou@hua.gr}}

\begin{document}
\maketitle

\begin{center}
    \centering
    \captionsetup{type=figure}
    \includegraphics[width=1\textwidth]{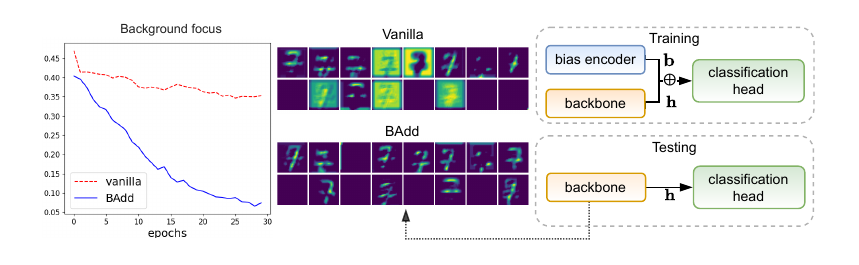}
    \captionof{figure}{
    During training on Biased-MNIST, where color-digit association is strong, a vanilla model struggles with bias mitigation as reducing reliance on the protected attribute increases loss for biased samples. Augmenting main features ($\mathbf{h}$) with protected attribute features ($\mathbf{b}$) allows for learning optimal filters without compromising loss. Consequently, the model learns fair representations of the digits, independent of color, as evidenced by the activation maps and mean activations on the background of samples, where bias occurs.
    }
    \label{fig:teaser}
\end{center}%

\begin{abstract}
Computer vision (CV) datasets often exhibit biases that are perpetuated by deep learning models. While recent efforts aim to mitigate these biases and foster fair representations, they fail in complex real-world scenarios. In particular, existing methods excel in controlled experiments involving benchmarks with single-attribute injected biases, but struggle with multi-attribute biases being present in well-established CV datasets. Here, we introduce BAdd, a simple yet effective method that allows for learning fair representations invariant to the attributes introducing bias by incorporating features representing these attributes into the backbone. BAdd is evaluated on seven benchmarks and exhibits competitive performance, surpassing state-of-the-art methods on both single- and multi-attribute benchmarks. Notably, BAdd achieves +27.5\% and +5.5\% absolute accuracy improvements on the challenging multi-attribute benchmarks, FB-Biased-MNIST and CelebA, respectively.
\end{abstract}

\section{Introduction}
Convolutional Neural Networks (CNNs) have demonstrated impressive capabilities, as evidenced by groundbreaking research across various Computer Vision (CV) tasks \cite{deng2019arcface,feichtenhofer2019slowfast,tan2020efficientdet}. 
However, a concerning issue has emerged alongside these advancements: the potential for bias in AI systems, disproportionately impacting specific groups \cite{barocasfairml, fabbrizzi2022survey, sarridis2023towards}.
Specifically, when AI systems base their decisions on attributes like age, gender, or race,
they become discriminatory. Considering the profound impact AI decisions can have on individuals' lives, such biases should be eliminated from any decision-making system \cite{bobadilla2013recommender,taigman2014deepface,creswell2018generative,tan2020efficientdet,sarridis2023towards}.

For AI systems processing visual data, bias often originates from the quality of the datasets used for training \cite{fabbrizzi2022survey}. 
One of the main types of bias they exhibit is the presence of stereotypes, which means that the datasets are constructed in such a way that specific groups of people are largely associated with certain attributes (e.g., women are illustrated wearing earrings in the majority of images). When such data is used to train Deep Learning (DL) models, these biases can act as ``shortcuts'', leading the model to prioritize irrelevant attributes in its decision-making process  \cite{zhang2018examining}. Motivated by this issue, sophisticated approaches have been proposed to enable learning fair representations that are independent of the biased attributes \cite{sarridis2023flac,hong2021bb,barbano2022fairkl}. 
Such methods often leverage labels associated with protected attributes to guide model training towards learning fair representations \cite{bahng2020rebias,cadene2019rubi,clark2019LM,hong2021bb,barbano2022fairkl,sarridis2023flac}. Techniques like adversarial training \cite{kim2019lnl,wang2019balanced}, regularization \cite{tartaglione2021end,hong2021bb}, and fairness-aware loss functions \cite{barbano2022fairkl,sarridis2023flac} are commonly employed. 
While these approaches aim to mitigate biases in established CV datasets, they largely rely on evaluation benchmarks that do not reflect the real-world complexities encountered in practice, as they typically comprise (i) uniformly distributed and artificially generated single-attribute biases, found in datasets like Biased-MNIST \cite{bahng2020rebias} and Corrupted-CIFAR10 \cite{hendrycks2018benchmarking}; 
(ii) enforced biases through data selection from established datasets, such as UTKFace \cite{zhifei2017cvpr} and CelebA \cite{hendrycks2018benchmarking}.
However, established CV datasets, such as  CelebA, exhibit vastly different properties in terms of biases:
\begin{itemize}
    \item Multiple attributes contribute to creating complex bias patterns. For instance, in CelebA, attributes like WearingEarrings, WearingLipstick, HeavyMakeup, and BlondHair are all highly associated with gender.
    \item The bias intensity also differs across different target classes. 
    In the example of CelebA, the BlondHair attribute has a 23.9\% and 2.1\% co-occurrence with females and males, respectively.
\end{itemize}
These fundamental differences between existing benchmarks and real-world datasets significantly complicate the task of mitigating bias, leading to significantly reduced performance of most existing approaches.

In this paper, we recognize the need to enhance the applicability of fairness-aware CV models in complex application settings and we introduce \METHOD, a simple yet effective methodology designed to mitigate bias stemming from single or multiple attributes effectively. Particularly, the proposed approach suggests that the injection of bias-capturing features into the penultimate layer's output enables learning representations invariant to the bias-capturing features (see Fig.~\ref{fig:teaser}).
At the core of \METHOD\ lies a core insight into the mechanism by which bias is introduced to the DL models during training via the minimization of the loss function. 
The model optimizes its parameters by taking advantage of biases present in the data, as doing so reduces the overall loss. Consequently, the model learns to prioritize features associated with the biased attributes, reinforcing and perpetuating the bias within its representations.
The intentional inclusion of bias-capturing features within the model's features ensures that the model can effectively encode the biases without them exerting undue influence on the loss function optimization. In essence, \METHOD\ decouples the learning of biased features from the optimization process and thus allows for learning fair representations. \METHOD\ outperforms state-of-the-art methods on a wide range of experiments involving four datasets with single attribute biases (i.e., Biased-MNIST, Biased-UTKFace, Waterbirds, and Corrupted-CIFAR10) and three datasets with multi-attribute biases (i.e., FB-Biased-MNIST, UrbanCars, and CelebA). Where \METHOD\ shines is on datasets involving multi-attribute biases, where it outperforms the state of the art by +27.5\% and +5.5\% absolute accuracy improvements on FB-Biased-MNIST and CelebA, respectively.
In summary, the paper makes the following contributions:
(i) we introduce \METHOD, an effective methodology for learning fair representations concerning one or more protected attributes by incorporating bias-capturing features into the model's representations (ii) providing an extensive evaluation involving seven benchmarks, showcasing the superiority of \METHOD\ on both single- and multi-attribute bias scenarios.

\section{Related Work}

\paragraph{Fairness-aware image benchmarks.}
Most standard benchmarks for evaluating bias mitigation techniques in CV involve single-attribute artificially generated biases. Biased-MNIST \cite{bahng2020rebias}, derived from the original MNIST dataset, associates each digit with a specific colored background. Similarly, Corrupted-CIFAR10 \cite{hendrycks2018benchmarking} introduces biased textures across its classes. The Waterbirds \cite{sagawa2019distributionally} dataset, constructed by cropping birds from the CUB-200 \cite{wah2011caltech} dataset and transferring them onto backgrounds from the Places dataset \cite{zhou2017places}, introduces biases through correlating bird species with certain backgrounds (i.e., habitat types).
On the other hand, datasets like Biased-UTKFace \cite{hong2021bb} and Biased-CelebA \cite{hong2021bb} are carefully selected subsets of UTKFace \cite{zhifei2017cvpr} and CelebA \cite{liu2015faceattributes}, respectively. These subsets are designed to showcase an association of 90\% between specific attributes, such as gender and race.
However, all these benchmarks share a crucial limitation: they are far from emulating the complexities of realistic scenarios, as they typically exhibit uniformly distributed single attribute biases.
To address this limitation, recent works focus on introducing benchmarks that involve multi-attribute biases, such as Biased-MNIST variations \cite{ahn2022mitigating, shrestha2022investigation, shrestha2022occamnets} and UrbanCars \cite{li2023whac}. Specifically, UrbanCars introduces a multi-attribute bias setting by incorporating biases related to both background and co-occurring objects and the task is to classify car body types into urban or country categories.

In addition to the above benchmarks, in this paper, we create a variation of Biased-MNIST, termed FB-Biased-MNIST, which builds on the bias introduced by background color in Biased-MNIST by injecting an additional bias through foreground color. Furthermore, we consider a benchmark that utilizes the original, unmodified CelebA dataset but focuses on evaluating performance against the most prominent bias-inducing attributes within the dataset. This allows for evaluating fairness-aware methodologies on multiple biases in a more realistic setting - without artificially injected biases.

\paragraph{Fairness-aware approaches.}

Efforts on learning fair representations using biased data encompass techniques like ensemble learning \cite{clark2019LM, wang2020DI}, contrastive learning \cite{hong2021bb, barbano2022fairkl}, adversarial frameworks \cite{xie2017controllable,alvi2018turning,kim2019lnl, song2019learning, wang2019balanced, adel2019one}, and regularization terms \cite{cadene2019rubi,tartaglione2021end,hong2021bb,sarridis2023flac}.
For instance, the Learning Not to Learn (LNL) approach \cite{kim2019lnl} penalizes models if they predict protected attributes, while the Domain-Independent (DI) approach \cite{wang2020DI} introduces the usage of domain-specific classifiers to mitigate bias. Entangling and Disentangling deep representations (EnD) \cite{tartaglione2021end} suggests a regularization term that entangles or disentangles feature vectors w.r.t. their target and protected attribute labels. FairKL \cite{barbano2022fairkl} and BiasContrastive-BiasBalance (BC-BB) \cite{hong2021bb} are contrastive learning-based approaches that try to mitigate bias by utilizing the pairwise similarities of the samples in the feature space. Finally, there are several works that can be employed without utilizing the protected attribute labels, such as Learned-Mixin (LM) \cite{clark2019LM}, Rubi \cite{cadene2019rubi}, ReBias \cite{bahng2020rebias}, Learning from Failure (LfF) \cite{nam2020LfF}, and FLAC \cite{sarridis2023flac}. The latter achieves state-of-the-art performance by utilizing a bias-capturing classifier and a sampling strategy that effectively focuses on the underrepresented groups. 
It is worth noting that methodologies for distributionally robust optimization \cite{sagawa2019distributionally,liu2021just,wu2023discover,qiu2023simple,li2022discover,li2023whac} are highly relevant, as they aim at mitigating biases arising from spurious correlations in the training data.
Recent efforts suggest approaches that are designed to be effective in scenarios involving multiple biases. For instance, OccamNets \cite{shrestha2022occamnets} propose modifying the network architecture to impose inductive biases that make the network robust to dataset bias. Furthermore, the Last Layer Ensemble (LLE) \cite{li2023whac} employs multiple augmentations to eliminate different biases (i.e., one type of augmentation for each type of bias). However, it requires extensive preprocessing (e.g., object segmentation), which makes it challenging or even infeasible to apply to different CV datasets. On the other hand, \METHOD\ is a simple yet effective approach that can be easily applied to any network architecture that processes images, as well as to any CV dataset.

\section{Methodology}
\subsection{Problem formulation}
Consider a dataset $\mathcal{D}$ comprising training samples $(\mathbf{x}^{(i)}, y^{(i)})$, where $\mathbf{x}^{(i)}$ represents the input sample and $y^{(i)}$ belongs to the set of target labels $\mathcal{Y}$. Let $h(\cdot)$ denote a model trained on $\mathcal{D}$ and $\mathbf{h}$ the model representation (e.g., output of penultimate model layer). Let also $\mathcal{T}$ be the domain of tuples of $q$ \textit{protected attributes} (e.g., the set $t = \left\{\text{\emph{male}}, 25, \text{\emph{black}}\right\}$ for protected attributes ``gender'', ``age'' and ``race''). The protected attributes should not be used to predict the targets in $\mathcal{Y}$, either because there is no causal relationship between them, or due to ethical or legal constraints. In addition, we consider a bias-capturing model $b(\cdot)$, which has been trained to predict the value of protected attribute(s) $t \in \mathcal{T}$ from $\mathbf{x}$, and the corresponding representation $\mathbf{b}$.

We define $\mathcal{D}$ as \textit{biased} with respect to the protected attributes in $\mathcal{T}$ if there is a significant imbalance in the co-occurrence of certain values in $\mathcal{Y}$ with a value or a combination of values of protected attributes in $\mathcal{T}$. 
Within a batch $\mathcal{B}$, samples exhibiting the dataset bias are termed bias-aligned ($\mathcal{A}$), while those that deviate from it are referred to as bias-conflicting ($\mathcal{C}$).
The set $\mathcal{D}$ is assumed to include at least some bias-conflicting examples.
Using such a biased dataset for training often introduces model bias, by leading $h$ to rely on features that encode information related to $t$. Thus, the objective is to mitigate these dependencies between representations $\mathbf{h}$ and $\mathbf{b}$, and consequently learn fair representations.

\subsection{Method description}
When training a model $h(\cdot)$ on a biased dataset $\mathcal{D}$, it often prioritizes learning features related to the protected attributes instead of features directly characterizing the target class. As already discussed, this phenomenon arises if there is a high correlation between protected attributes and targets and if the protected attribute's visual characteristics are easier to capture than the visual characteristics of the target \cite{zhang2018examining}. 
Below, we delve into the details behind a vanilla model's limited ability to mitigate this kind of bias and explain how the proposed approach addresses this limitation.

First, let us consider the Cross-Entropy loss on a batch of samples $\mathcal{B}=\mathcal{A}\cup\mathcal{C}$:
\begin{equation}
\label{eq:loss}
\begin{split}
    \mathcal{L} = &-\frac{1}{N}\sum_{i\in\mathcal{B}}\sum_{k=1}^{K}y_k^{(i)}\log \hat{y}_k^{(i)} = \\
    &-\frac{1}{N}\sum_{i\in\mathcal{A}}\sum_{k=1}^{K}y_k^{(i)}\log \hat{y}_k^{(i)}-\frac{1}{N}\sum_{i\in\mathcal{C}}\sum_{k=1}^{K}y_k^{(i)}\log \hat{y}_k^{(i)}=\\
    &\mathcal{L}_{\mathcal{A}} + \mathcal{L}_{\mathcal{C}},
\end{split}
\end{equation}
where 
$N$ is the number of samples within a batch, and $K$ the number of target classes. The predictions $\hat{y}_k^{(j)}$ are computed as follows:
\begin{equation}
    \hat{y}_k^{(j)} = \sigma_k(\mathbf{z}(\mathbf{x}^{(j)};\boldsymbol{\theta})),
\end{equation}
where $j$ the index of input sample $\mathbf{x}^{(j)}$, $\boldsymbol{\theta}$ the learnable parameters of model $h(\cdot)$ and $\sigma_k$ the $k$-th class probability after applying the softmax function on the logits $\mathbf{z}(\mathbf{x}^{(j)}; \boldsymbol{\theta})$. 

\begin{figure}
    \centering
    \includegraphics[width=0.55\linewidth]{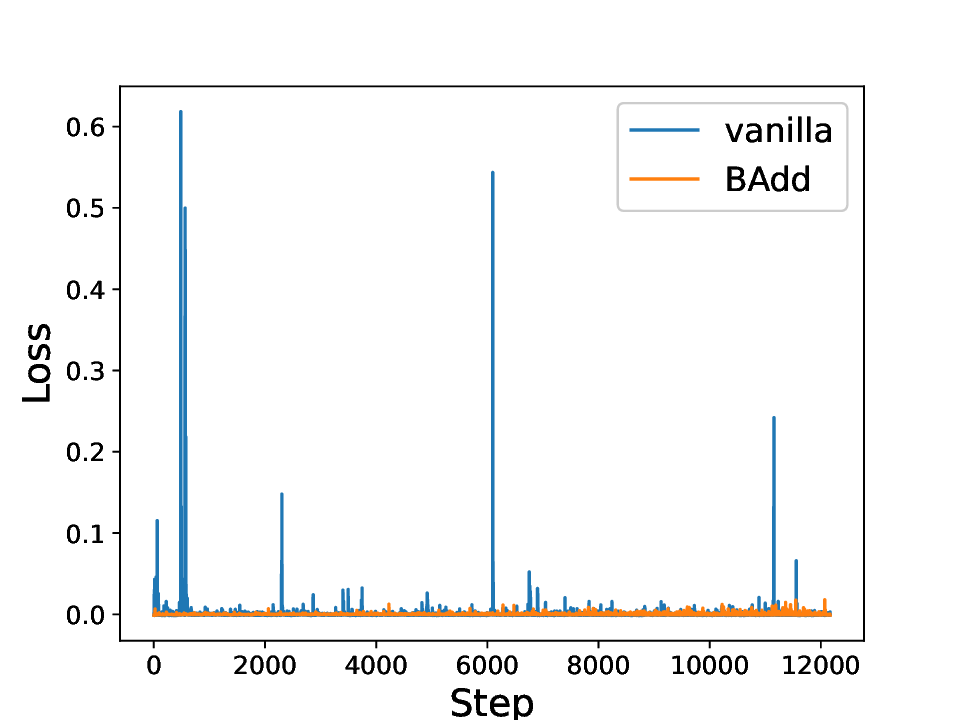}
    \caption{Vanilla vs \METHOD : Bias-aligned samples loss on Biased-MNIST dataset.}
    \label{fig:spikes}
\end{figure}

Given that $||\mathcal{A}||>>||\mathcal{C}||$, we can  assume that there exists a point in the training process at which
the model 
has 
learned to be accurate on the bias-aligned samples $\mathcal{A}$ misguidedly relying on protected attributes' features, so that 
$\mathcal{L}_{\mathcal{A}}\approx0$, while at the same time 
$\mathcal{L}_{\mathcal{C}}>>0$.
Consequently, backpropagating the gradients of $\mathcal{L}$ 
will update the parameters $\boldsymbol{\theta}$ in a way that steers the model towards accurately predicting the bias-conflicting samples $\mathcal{C}$ in order to further reduce $\mathcal{L}$, which stops reliance of $h(\cdot)$ on the protected attributes.
The major limitation of a vanilla model is directly connected to the loss behavior when the model processes the next mini-batch. In particular, the step the model makes towards correctly predicting the samples in $\mathcal{C}$, thus reducing $\mathcal{L}_{\mathcal{C}}$, severely affects the loss w.r.t. the bias-aligned samples, which is now $\mathcal{L}_{\mathcal{A}}>>0$, as $h(\cdot)$ relies less 
on the protected attributes and at the same time it is impossible to learn to encode the target with only one batch of $||C||$ bias-conflicting samples. This leads to a loss spike for the bias-aligned samples that in the next iteration will restore the model's parameters $\boldsymbol{\theta}$ to their initial state (encoding the protected attributes' features) in order to again achieve a much lower $\mathcal{L}$. 
Figure~\ref{fig:spikes} illustrates the loss w.r.t. the bias-aligned samples for approximately 12k training steps. As one may observe, the vanilla model tends to present these loss spikes that prevent the model from mitigating the bias. 

To support these arguments in more concrete terms, let us consider the derivative of the loss of Eq.~\eqref{eq:loss} with respect to a parameter $\theta_0$ for the $i$-th sample:

\begin{equation}
    \begin{split}
    \frac{\partial\mathcal{L}^{(i)}}{\partial\theta_0} = & y_{\kappa} \frac{\partial\log \sigma_{\kappa}(\mathbf{z}(\mathbf{x}^{(i)};\boldsymbol{\theta}))}{\partial\theta_0}= \\ &y_{\kappa}\frac{1}{\sigma_{\kappa}(\mathbf{z}(\mathbf{x}^{(i)};\boldsymbol{\theta}))}\frac{\partial\sigma_{\kappa}(\mathbf{z}(\mathbf{x}^{(i)};\boldsymbol{\theta}))}{\partial\theta_0},
    \end{split}
\end{equation}
where $\kappa$ is the correct class, according to the ground truth (i.e., $y_{\kappa}=1$). Setting $A_0^{(i)}= \frac{\partial\sigma_{\kappa}(\mathbf{z}(\mathbf{x}^{(i)};\boldsymbol{\theta}))}{\partial\theta_0}$ the derivative for a batch becomes
\begin{equation}
\label{eq:loss_derivative}
    \frac{\partial\mathcal{L}}{\partial\theta_0} = - \frac{1}{N}\Big(\sum_{i: \mathbf{x}^{(i)}\in\mathcal{A}}\frac{1}{\sigma_{\kappa}^{(i)}}A_0^{(i)}+ \sum_{j: \mathbf{x}^{(j)}\in\mathcal{C}}\frac{1}{\sigma_{\kappa}^{(j)}}A_0^{(j)} \Big)
\end{equation}
After the model has learned to predict the targets based on the protected attributes, $\sigma_{\kappa}^{(i)}$ is large (close to 1) while $A_0^{(i)}$ is small, as $h(\cdot)$ already correctly predicts samples in $\mathcal{A}$. In contrast, $\sigma_{\kappa}^{(j)}$ is small while $A_0^{(j)}$ is large. The model update therefore strongly depends on the samples in $\mathcal{C}$. After the update step, however, $\sigma_{\kappa}^{(i)}$ becomes smaller,  $A_{0}^{(i)}$ becomes larger and given that $||\mathcal{A}||>>||\mathcal{C}||$, the derivative is now dominated by samples in $\mathcal{A}$, and the parameters revert back to their previous values. In other words, any progress the model makes towards reducing its bias is counteracted by the loss function, which is lower when the model focuses on the easier-to-learn, biased samples. This essentially traps the model in a vicious cycle being condemned to encode the protected attributes instead of the targets.


To address this limitation, \METHOD\ proposes combining the features capturing the protected attribute, represented by $\mathbf{b}$, with the main model's representation, represented by $\mathbf{h}$. This combined representation $\mathbf{h}+\mathbf{b}$ is then fed to the final classification layer.\footnote{The $\mathbf{b}$ features can be obtained either by using a bias-capturing classifier (i.e., protected attribute extractor) or by using a regressor that encodes the protected attribute labels into a feature vector matching the dimension of  $\mathbf{h}$.
The former is considered less prone to overfitting due to its diversity in the feature space, while the latter is easier to implement as it does not require any additional training procedure.} 
Thus, during training, model predictions are computed as
\begin{equation}
   \hat{y}_k^{(j)} = \sigma_k(\mathbf{W}(\mathbf{h}(\mathbf{x}^{(j)}; \boldsymbol{\theta}) + \mathbf{b}(\mathbf{x}^{(j)})) + \boldsymbol{\rho}),
\end{equation}
where $\mathbf{W}$ and $\boldsymbol{\rho}$ are the parameters of the last linear layer of $h(\cdot)$.
By incorporating the protected attribute features $\mathbf{b}$ into the training, we equip the model with the necessary information to consistently predict accurately the bias-aligned samples.
This means that the $\mathcal{L}_{\mathcal{A}}$ values are consistently close to 0, preventing the loss spikes (see Fig.~\ref{fig:spikes}), and thus enabling the $\mathbf{h}$ to encode the targets rather than the protected attributes, without having a negative impact on the loss of the bias-aligned samples. In terms of the training process implied by Eq. \eqref{eq:loss_derivative}, the addition of $\mathbf{b}$ entails invariably large $\sigma_{\kappa}^{(i)}$ and small $A_0^{(i)}$, thus forcing the model updates to depend on the samples of $\mathcal{C}$ consequently eliminating the effect of bias-aligned samples. 
After training $h(\cdot)$, fine-tuning the classification layer with only the $\mathbf{h}$ features is required, as $\mathbf{b}$ should not be involved in inference, i.e. the predictions are derived using the following:
\begin{equation}
   \hat{y}_k^{(j)} = \sigma_k(\mathbf{W}\mathbf{h}(\mathbf{x}^{(j)}; \boldsymbol{\theta}) + \boldsymbol{\rho}),
\end{equation}


\section{Experimental setup}
\subsection{Datasets}
\paragraph{Single-attribute bias.} 
Biased-MNIST \cite{bahng2020rebias}, derived from the original MNIST dataset \cite{lecun1998mnist}, serves as a benchmark for assessing bias mitigation methods. It features digits with colored backgrounds, introducing bias through the association of each digit with a specific color. The degree of bias, represented by the probability $q$, determines the strength of this association. We consider four variations of Biased-MNIST with $q$ values of 0.99, 0.995, 0.997, and 0.999, as commonly used in previous works.
Biased-CelebA \cite{hong2021bb} is a subset of the well-known CelebA dataset consisting of facial images annotated with 40 binary attributes, which considers gender as the protected attribute, while HeavyMakeup and BlondHair serve as the target labels. Similarly, the Biased-UTKFace \cite{hong2021bb} dataset is a subset of the UTKFace dataset comprising facial images annotated with gender, race, and age labels. Gender is the target label, with race or age considered as protected attributes. In both Biased-CelebA and Biased-UTKFace, the enforced correlation between the target and protected attributes is 90\%.
As for the Corrupted-CIFAR10 dataset introduced in \cite{hendrycks2018benchmarking}, it consists of 10 classes with texture-related biases uniformly distributed in the training data. This dataset provides four correlation ratios: 0.95, 0.98, 0.99, and 0.995, offering a range of bias strengths for evaluation purposes. Finally, the Waterbirds dataset demonstrates a co-occurrence of 95\% between waterbirds (or landbirds) and aquatic environments (or terrestrial environments) as background.
\begin{table}[t]
\begin{center}
 \caption{Fairness of a vanilla gender classifier trained on default CelebA w.r.t. potentially biased attributes.}
    \label{tab:celeba_attrs}
 \begin{tabular}{lcc} 
 \toprule
 \multirow{ 2}{*}{Attribute}  &    \multicolumn{2}{c}{Accuracy}     \\ \cline{2-3}
          & Unbiased & Bias-conflicting \\ \midrule
Smiling & 98.6 & 98.5 \\
WearingNecklace & 98.1 & 97.3 \\
WearingEarrings & 97.7 & 96.3 \\
BlondHair & 96.9 & 94.9 \\
Eyeglasses & 96.5 & 94.5 \\
WearingLipstick & 95.2 & 91.1 \\
HeavyMakeup & 93.0 & 86.7 \\
\bottomrule
\end{tabular}
\end{center}
\end{table}
\paragraph{Multi-attribute bias.} Similar to the Biased-MNIST, we create FB-Biased-MNIST, an extension that enhances the bias introduced by the background color in Biased-MNIST. Specifically, FB-Biased-MNIST introduces an additional layer of bias by injecting foreground color biases into the dataset. Considering the increased complexity of this dataset compared to the Biased-MNIST, we opt for lower $q$ values, namely 0.9, 0.95, and 0.99. Furthermore, the UrbanCars dataset is an artificially generated dataset that shows a 95\% co-occurrence between car body types and both the background and certain objects relevant to urban or rural regions.
We also assess the performance of bias mitigation methods on established CV datasets devoid of injected biases, whether artificially generated or introduced through subset selection. To this end, we utilize the default CelebA \cite{liu2015faceattributes} dataset. To properly select the attributes introducing  bias, we consider the performance disparities of a standard gender classifier trained on CelebA with respect to various potentially biased attributes. Subsequently, we identify the top two attributes with most significant impact on the model's performance. As shown in Tab.~\ref{tab:celeba_attrs}, WearingLipstick and HeavyMakeup are those two attributes. 
\begin{table}[t]
\begin{center}
 \caption{CelebA gender bias degree $q$ w.r.t. the attributes WearingLipstick and HeavyMakeup.}
    \label{tab:celeba_q}
 \begin{tabular}{lcc} 
 \toprule
 \multirow{ 2}{*}{Attribute}  &    \multicolumn{2}{c}{$q$}     \\ \cline{2-3}
          & Females & Males \\ \midrule
WearingLipstick & 0.806 & 0.994 \\
HeavyMakeup  & 0.663 & 0.997 \\
\bottomrule
\end{tabular}
\end{center}
\end{table}

\subsection{Model architecture}
In experiments involving the MNIST-based datasets, namely Biased-MNIST and FB-Biased-MNIST, we utilize a simple CNN architecture outlined in \cite{bahng2020rebias}, which comprises four convolutional layers with 7$\times$7 kernels and a classification head. For the experiments involving the Biased-UTKFace, Corrupted-CIFAR10, and CelebA datasets, we adopt the ResNet-18 architecture \cite{he2016deep}. For Waterbirds and UrbanCars datasets, we use Resnet-50 networks. 

\subsection{Implementation details and evaluation protocol}
We employ the Adam optimizer with a 0.001 initial learning rate, which is divided by 10 every 1/3 of the training epochs. Batch size is fixed at 128 and weight decay is set to $10^{-4}$.
Following previous works \cite{hong2021bb,sagawa2019distributionally,li2023whac}, we train the models on Biased-MNIST and FB-Biased-MNIST datasets for 80 epochs. For Biased-UTKFace and CelebA,  training duration is set to 20 and 40 epochs, respectively. As for the Corrupted-CIFAR10 dataset, models are trained for 100 epochs using a cosine annealing scheduler. For the Waterbirds and UrbanCars datasets, we do not use a learning rate scheduler, and the models are trained for 300 and 100 epochs, respectively.
Following the initial training phase, the classification head of all models is fine-tuned for an additional 20 epochs.
All experiments were conducted on a single NVIDIA RTX-3090 Ti GPU and repeated for 5 different random seeds.
Regarding the evaluation protocol, the test sets used for Biased-MNIST and FB-Biased-MNIST have $q=0.1$ to ensure unbiased evaluation. For Biased-UTKFace and CelebA datasets, we utilize both bias-conflicting and unbiased accuracy as in \cite{sarridis2023flac,hong2021bb}. The official unbiased test set is utilized for Corrupted-CIFAR10. For the Waterbirds, we employ the average accuracy between different groups and the Worst-Group (WG) accuracy. Finally, for the UrbanCars dataset, we measure the In Distribution Accuracy (I.D. Acc) which is the
weighted average accuracy w.r.t. the different groups, where the correlation ratios are the weights. The I.D. Acc is used as a baseline to measure the accuracy drop with respect to the background (BG Gap), co-occurring objects (CoObj Gap), and both the background and co-occurring objects (BG+CoObj Gap).



\begin{figure}[t]
    \centering
    \begin{subfigure}{0.45\linewidth}
        \centering
        \includegraphics[width=\linewidth]{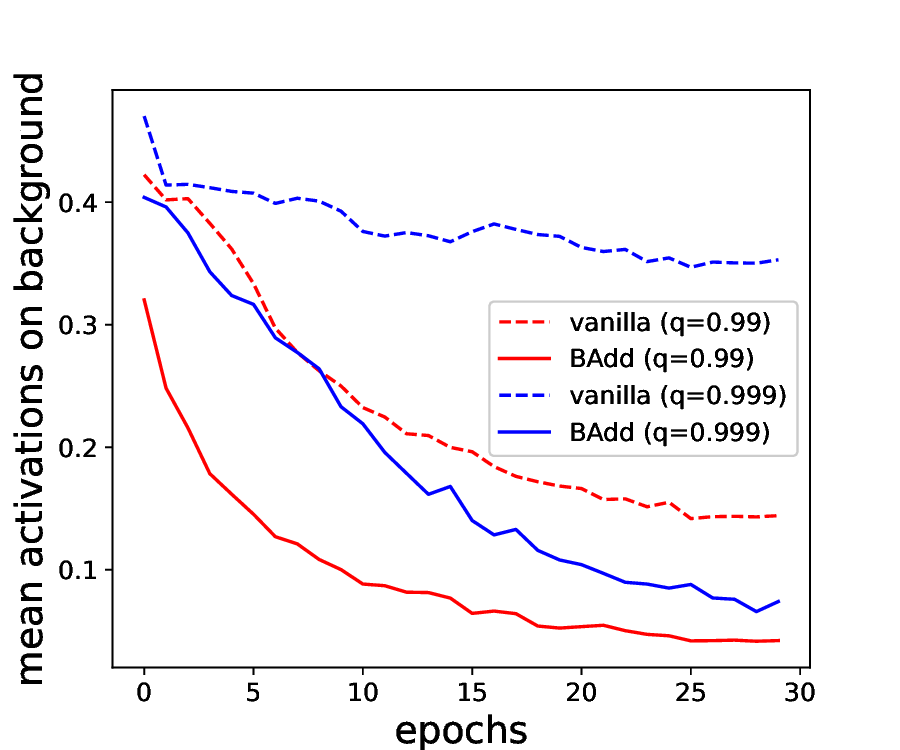}
        \caption{Mean activation values of the first convolutional layer on sample backgrounds.}
    \end{subfigure}
    \begin{subfigure}{0.45\linewidth}
        \centering
        \includegraphics[width=\linewidth]{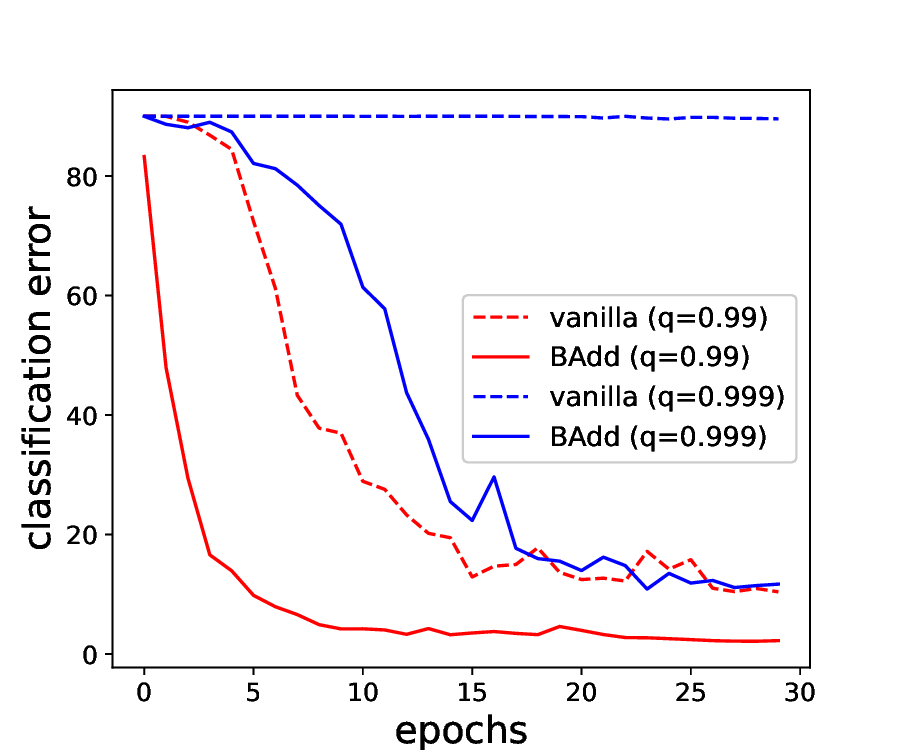}
        \caption{Classification error during the first 30 training epochs.}
        \label{subfig:acc}
    \end{subfigure}
    
    \caption{Comparison of mean biased filter activation values and classification error between Vanilla and \METHOD\ on Biased-MNIST.}
    \label{fig:act}
\end{figure}

\section{Results}
\subsection{Single attribute biases}
Table \ref{tab:mnist} showcases the performance of \METHOD\ against nine competing methods. As one may observe, the proposed approach consistently surpasses state-of-the-art, demonstrating accuracy improvements ranging from 0.1\% to 0.8\% across different $q$ values. 
Furthermore, Fig.~\ref{fig:act} illustrates the mean activations in image regions where bias occurs alongside the corresponding classification errors for both the Vanilla and \METHOD\ approaches. As one may observe, the proposed method effectively reduces activations in areas where bias appears, leading to significant improvements in classification performance. This effect is particularly pronounced in experiments with $q=0.999$, where the Vanilla approach struggles to mitigate the impact of the biased attribute. Furthermore, the efficacy of \METHOD\ to learn representations that are independent of the protected attribute is illustrated in Tab.~\ref{tab:sim}. Specifically, Tab.~\ref{tab:sim} shows the mean pairwise cosine similarity between 10 variations of each Biased-MNIST test sample, where each sample variation has a different background color (i.e., protected attribute). As one may observe, \METHOD\ showcases similarity values close to 1 for all correlation ratios, which is not the case for the vanilla model that cannot maintain high similarities when the correlation ratio increases (e.g., 0.416 similarity for $q=0.999$).

\begin{table}[b]
\centering
\caption{Evaluation on Biased-MNIST for different bias levels. }
\label{tab:mnist}
\begin{tabular}{lcccc} 
\toprule
\multirow{2}{*}{Method} & \multicolumn{4}{c}{$q$} \\ 
\cmidrule(lr){2-5}
& 0.99 & 0.995 & 0.997 & 0.999 \\ 
\midrule
Vanilla & 90.8\scriptsize{$\pm$0.3} & 79.5\scriptsize{$\pm$0.1} & 62.5\scriptsize{$\pm$2.9} & 11.8\scriptsize{$\pm$0.7} \\ 
LM \cite{clark2019LM} & 91.5\scriptsize{$\pm$0.4} & 80.9\scriptsize{$\pm$0.9} & 56.0\scriptsize{$\pm$4.3} & 10.5\scriptsize{$\pm$0.6}\\
Rubi \cite{cadene2019rubi} & 85.9\scriptsize{$\pm$0.1} & 71.8\scriptsize{$\pm$0.5} & 49.6\scriptsize{$\pm$1.5} & 10.6\scriptsize{$\pm$0.5}\\
ReBias \cite{bahng2020rebias} & 88.4\scriptsize{$\pm$0.6} & 75.4\scriptsize{$\pm$1.0} & 65.8\scriptsize{$\pm$0.3} & 26.5\scriptsize{$\pm$1.4}\\
LfF \cite{nam2020LfF} & 95.1\scriptsize{$\pm$0.1} & 90.3\scriptsize{$\pm$1.4} & 63.7\scriptsize{$\pm$20.3} & 15.3\scriptsize{$\pm$2.9}\\
LNL \cite{kim2019lnl} & 86.0\scriptsize{$\pm$0.2} & 72.5\scriptsize{$\pm$0.9} & 57.2\scriptsize{$\pm$2.2} & 18.2\scriptsize{$\pm$1.2}\\
EnD \cite{tartaglione2021end} & 94.8\scriptsize{$\pm$0.3} & 94.0\scriptsize{$\pm$0.6} & 82.7\scriptsize{$\pm$0.3} & 59.5\scriptsize{$\pm$2.3}\\         
BC-BB \cite{hong2021bb} & 95.0\scriptsize{$\pm$0.9} & 88.2\scriptsize{$\pm$2.3} & 82.8\scriptsize{$\pm$4.2} & 30.3\scriptsize{$\pm$11.1} \\
FairKL \cite{barbano2022fairkl} & 97.9\scriptsize{$\pm$0.0} & 97.0\scriptsize{$\pm$0.0} & 96.2\scriptsize{$\pm$0.2} & 90.5\scriptsize{$\pm$1.5}\\ 
FLAC \cite{sarridis2023flac} & 97.9\scriptsize{$\pm$0.1} & 96.8\scriptsize{$\pm$0.0} & 95.8\scriptsize{$\pm$0.2} & 89.4\scriptsize{$\pm$0.8}\\ 
\midrule         
\METHOD & \textbf{98.1\scriptsize{$\pm$0.2}} & \textbf{97.3\scriptsize{$\pm$0.2}} & \textbf{96.3\scriptsize{$\pm$0.2}} & \textbf{91.7\scriptsize{$\pm$0.6}} \\
\bottomrule
\end{tabular}
\end{table}
\begin{table}[t]
\begin{center}
\caption{Mean pairwise cosine similarity between 10 variations of each Biased-MNIST test sample, where each sample variation has a different background color.}\label{tab:sim}
    \begin{tabular}{lcccc} 
    \toprule
          \multirow{ 2}{*}{Method}  &    \multicolumn{4}{c}{$q$}     \\ \cline{2-5}
 & 0.99 & 0.995 & 0.997 & 0.999 \\ \midrule
Vanilla   & 0.889 & 0.854 & 0.811 & 0.416 \\ 
\METHOD & 0.985 & 0.985 & 0.980 & 0.973      \\ 
\bottomrule
\end{tabular}
\end{center}
\end{table}

Table \ref{tab:utk} illustrates the performance comparison of \METHOD\ against the competing methods on the Biased-UTKFace dataset, where race and age are considered as protected attributes. Across both protected attributes, the proposed approach outperforms the second best-performing competing methods on \textit{bias-conflicting} samples, achieving improvements of +1.1\% and +1.9\%, respectively. For the \textit{unbiased} performance, \METHOD\ exhibits only marginal differences compared to the state-of-the-art methods, with increases of 0.2\% and decreases of 0.3\%, respectively. 


\begin{table}[t]
\centering
\caption{Evaluation of the proposed method on Biased-UTKFace for two different protected attributes, namely \textit{race} and \textit{age}, with \textit{gender} as the target attribute.}
\label{tab:utk}
\begin{tabular}{lcccc} 
\toprule
 & \multicolumn{4}{c}{Bias} \\ 
\cmidrule(lr){2-5}
Method & \multicolumn{2}{c}{Race} & \multicolumn{2}{c}{Age} \\ 
\cmidrule(lr){2-3} \cmidrule(lr){4-5}
& Unbiased & Bias-conflicting & Unbiased & Bias-conflicting \\ 
\midrule
Vanilla & 87.4\scriptsize{$\pm$0.3} & 79.1\scriptsize{$\pm$0.3} & 72.3\scriptsize{$\pm$0.3} & 46.5\scriptsize{$\pm$0.2} \\ 
LNL \cite{kim2019lnl} & 87.3\scriptsize{$\pm$0.3} & 78.8\scriptsize{$\pm$0.6} & 72.9\scriptsize{$\pm$0.1} & 47.0\scriptsize{$\pm$0.1} \\ 
EnD \cite{tartaglione2021end} & 88.4\scriptsize{$\pm$0.3} & 81.6\scriptsize{$\pm$0.3} & 73.2\scriptsize{$\pm$0.3} & 47.9\scriptsize{$\pm$0.6} \\ 
BC-BB \cite{hong2021bb} & 91.0\scriptsize{$\pm$0.2} & 89.2\scriptsize{$\pm$0.1} & 79.1\scriptsize{$\pm$0.3} & 71.7\scriptsize{$\pm$0.8} \\ 
FairKL \cite{barbano2022fairkl} & 85.5\scriptsize{$\pm$0.7} & 80.4\scriptsize{$\pm$1.0} & 72.7\scriptsize{$\pm$0.2} & 48.6\scriptsize{$\pm$0.6} \\
FLAC \cite{sarridis2023flac} & 92.0\scriptsize{$\pm$0.2} & 92.2\scriptsize{$\pm$0.7} & \textbf{80.6\scriptsize{$\pm$0.7}} & 71.6\scriptsize{$\pm$2.6} \\ 
\midrule         
\METHOD & \textbf{92.2\scriptsize{$\pm$0.2}} & \textbf{93.3\scriptsize{$\pm$0.2}} & 80.3\scriptsize{$\pm$0.8} & \textbf{73.6\scriptsize{$\pm$1.0}} \\
\bottomrule
\end{tabular}
\end{table}

In the final single-attribute evaluation scenario, biases stemming from image background or textures are considered. As for the texture biases, the results obtained on the Corrupted-CIFAR10 dataset for four different bias ratios are summarized in Tab.~\ref{tab:cifar}. Given the complexity of training a bias-capturing classifier in this scenario, \METHOD\ is implemented using a linear regressor to obtain feature vectors of the desired size from one-hot vectors representing the texture labels. Notably, even without utilizing a bias-capturing classifier, \METHOD\ consistently outperforms state-of-the-art across all Corrupted-CIFAR10 variations. Specifically, it achieves improvements of 6.5\%, 3.1\%, 3.4\%, and 1.6\% for correlation ratios of 0.95, 0.98, 0.99, and 0.995, respectively. Table~\ref{tab:waterbirds}, demonstrates the performance of \METHOD\ on the Waterbirds dataset compared to the state-of-the-art methods for distributionally robust optimization. Here, \METHOD\ reaches the state-of-the-art WG accuracy, i.e., 92.9\%, and demonstrates competitive average accuracy, i.e., 93.6\%. 

         
         
         


\begin{table}[h]
\centering
\caption{Evaluation on Corrupted-Cifar10.}
\label{tab:cifar}
\begin{tabular}{lcccc} 
\toprule
\multirow{2}{*}{Method} & \multicolumn{4}{c}{$q$} \\ 
\cmidrule(lr){2-5}
& 0.95 & 0.98 & 0.99 & 0.995 \\ 
\midrule
Vanilla & 39.4\scriptsize{$\pm$0.6} & 30.1\scriptsize{$\pm$0.7} & 25.8\scriptsize{$\pm$0.3} & 23.1\scriptsize{$\pm$1.2}\\ 
EnD  \cite{tartaglione2021end} & 36.6\scriptsize{$\pm$4.0} & 34.1\scriptsize{$\pm$4.8} & 23.1\scriptsize{$\pm$1.1} & 19.4\scriptsize{$\pm$1.4}\\
ReBias \cite{bahng2020rebias}  & 43.4\scriptsize{$\pm$0.4} & 31.7\scriptsize{$\pm$0.4} & 25.7\scriptsize{$\pm$0.2} & 22.3\scriptsize{$\pm$0.4}\\
LfF  \cite{nam2020LfF}& {50.3}\scriptsize{$\pm$1.6} & {39.9}\scriptsize{$\pm$0.3} & {33.1}\scriptsize{$\pm$0.8} & 28.6\scriptsize{$\pm$1.3}\\ 
FairKL \cite{barbano2022fairkl} & 50.7\scriptsize{$\pm$0.9} & 41.5\scriptsize{$\pm$0.4} & 36.5\scriptsize{$\pm$0.4} & 33.3\scriptsize{$\pm$0.4}\\
FLAC \cite{sarridis2023flac} & 53.0\scriptsize{$\pm$0.7}  & 46.0\scriptsize{$\pm$0.2} & 39.3\scriptsize{$\pm$0.4} & 34.1\scriptsize{$\pm$0.5}\\ 
\midrule         
\METHOD & \textbf{59.5\scriptsize{$\pm$0.5}} & \textbf{49.1\scriptsize{$\pm$0.3}} & \textbf{42.7\scriptsize{$\pm$0.2}} & \textbf{35.7\scriptsize{$\pm$0.6}} \\
\bottomrule
\end{tabular}
\end{table}

\begin{table}[h]
\centering
\caption{Evaluation on Waterbirds.}
\label{tab:waterbirds}
\begin{tabular}{lcccc} 
\toprule
Method & WG Acc. & Avg. Acc.\\ 
\midrule       
JTT \cite{liu2021just}& 86.7\scriptsize{$\pm$1.5} & 93.3\scriptsize{$\pm$0.3} \\
DISC \cite{wu2023discover}& 88.7\scriptsize{$\pm$0.4} & 93.8\scriptsize{$\pm$0.7} \\
GroupDro \cite{sagawa2019distributionally} & 90.6\scriptsize{$\pm$1.1} & 91.8\scriptsize{$\pm$0.3} \\

DFR \cite{qiu2023simple}& \textbf{92.9\scriptsize{$\pm$0.2}} & \textbf{94.2\scriptsize{$\pm$0.4}} \\
\midrule
\METHOD & \textbf{92.9\scriptsize{$\pm$0.3}} & 93.6\scriptsize{$\pm$0.2}  \\
\bottomrule
\end{tabular}
\end{table}

\subsection{Multi-attribute biases}
As previously discussed, evaluating bias mitigation performance solely in single-attribute scenarios provides an initial assessment but fails to capture the complexities of real-world settings. In this section, we present the performance of \METHOD\ in two multi-attribute bias evaluation setups, namely on FB-Biased-MNIST and CelebA datasets. As depicted in Tab.~\ref{tab:multimnist}, competing methods struggle to effectively mitigate bias on the FB-Biased-MNIST dataset, while \METHOD\ consistently outperforms the second-best performing methods by significant margins of 8\%, 23\%, and 27.5\% for correlation ratios of 0.9, 0.95, and 0.99, respectively. Notably, even in an artificial dataset like FB-Biased-MNIST, existing approaches struggle to address multiple biases, underscoring the necessity for methodologies applicable to complex settings, such as \METHOD. 
Table~\ref{tab:urbancars} demonstrates the performance of \METHOD\ on UrbanCars, a dataset with artificially injected bias that is much more challenging than FB-Biased-MNIST. 
As observed, most compared methods struggle to address both the background and the co-occurring object biases. The only exception is LLE, which employs architectural modifications and specific bias-oriented augmentations to tackle each type of bias. However, it should be stressed that this approach requires extensive preprocessing, such as object segmentation, making its application to other CV datasets challenging or even infeasible.
Finally, an example of a real-world dataset - without artificially injected biases - is the default CelebA dataset, which exhibits multiple biases. As shown in Tab.~\ref{tab:celeba}, \METHOD\ consistently improves fairness performance for both biased attributes, achieving absolute accuracy improvements of +3.5\% and +5.5\% for the bias-conflicting samples and +1.1\% and +2.1\% for the unbiased sets compared to the second-best performing methods.
         


\begin{table}[h]
\centering
\caption{Evaluation on FB-Biased-MNIST for different bias levels. }
\label{tab:multimnist}
\begin{tabular}{lccc} 
\toprule
\multirow{2}{*}{Method} & \multicolumn{3}{c}{$q$} \\ 
\cmidrule(lr){2-4}
 & 0.9 & 0.95 & 0.99\\ 
\midrule
Vanilla & 82.5\scriptsize{$\pm$0.8} & 57.9\scriptsize{$\pm$1.7} & 25.5\scriptsize{$\pm$0.6} \\ 
EnD \cite{tartaglione2021end} & 82.5\scriptsize{$\pm$1.0} & 57.5\scriptsize{$\pm$2.0} & 25.7\scriptsize{$\pm$0.8} \\
BC-BB \cite{hong2021bb}  & 80.9\scriptsize{$\pm$2.4} & 66.0\scriptsize{$\pm$2.4} & 40.9\scriptsize{$\pm$3.4} \\
FairKL \cite{barbano2022fairkl} & 87.6\scriptsize{$\pm$0.8} & 61.6\scriptsize{$\pm$2.6} & 42.0\scriptsize{$\pm$1.1}\\ 
FLAC \cite{sarridis2023flac} & 84.4\scriptsize{$\pm$0.8} & 63.1\scriptsize{$\pm$1.7} & 32.4\scriptsize{$\pm$1.1} \\ 
\midrule         
\METHOD & \textbf{95.6\scriptsize{$\pm$0.3}} & \textbf{89.0\scriptsize{$\pm$1.8}} & \textbf{69.5\scriptsize{$\pm$2.5}} \\
\bottomrule
\end{tabular}
\end{table}

\begin{table}[h]
\centering
\caption{Evaluation on UrbanCars.}
\label{tab:urbancars}
\begin{tabular}{lcccccc} 
\toprule
Method & I.D. Acc & BG Gap & CoObj Gap & BG+CoObj Gap\\ 
\midrule       
LfF \cite{nam2020LfF} & 97.2 & -11.6  &  -18.4 & -63.2 \\
JTT \cite{liu2021just} & 95.9 & -8.1  &  -13.3 & -40.1 \\
Debian \cite{li2022discover} & 98.0 & -14.9  &  -10.5 & -69.0 \\
GroupDro \cite{sagawa2019distributionally} & 91.6 & -10.9  &  -3.6 & -16.4 \\
DFR \cite{qiu2023simple} & 89.7 & -10.7  &  -6.9 & -45.2 \\
LLE \cite{li2023whac} & 96.7 & \textbf{-2.1}  &  -2.7 & -5.9 \\
\midrule
\METHOD & 91.0\scriptsize{$\pm$0.7} & -4.3\scriptsize{$\pm$0.4} &  \textbf{-1.6\scriptsize{$\pm$1.0}} &\textbf{-3.9\scriptsize{$\pm$0.4}} \\
\bottomrule
\end{tabular}
\end{table}


\begin{table}[h]
\centering
\caption{Evaluation of the proposed method on CelebA for multiple attributes introducing bias, namely \textit{WearingLipstick} and \textit{HeavyMakeup}. \textit{Gender} is the target attribute.}
\label{tab:celeba}
\begin{tabular}{lcccc} 
\toprule
 & \multicolumn{4}{c}{Biases} \\ 
\cmidrule(lr){2-5}
Method & \multicolumn{2}{c}{WearingLipstick} & \multicolumn{2}{c}{HeavyMakeup} \\ 
\cmidrule(lr){2-3} \cmidrule(lr){4-5}
& Unbiased & Bias-conflicting & Unbiased & Bias-conflicting \\ 
\midrule
Vanilla & 95.2\scriptsize{$\pm$0.3} & 91.1\scriptsize{$\pm$0.6} & 93.0\scriptsize{$\pm$0.8} & 86.7\scriptsize{$\pm$1.6} \\ 
EnD \cite{tartaglione2021end} & 95.1\scriptsize{$\pm$0.4} & 91.0\scriptsize{$\pm$0.7} & 92.3\scriptsize{$\pm$0.7} & 85.3\scriptsize{$\pm$1.5} \\ 
BC-BB \cite{hong2021bb} & 91.6\scriptsize{$\pm$2.6} & 85.8\scriptsize{$\pm$5.1} & 89.7\scriptsize{$\pm$2.3} & 81.8\scriptsize{$\pm$4.5} \\ 
FairKL \cite{barbano2022fairkl} & 82.7\scriptsize{$\pm$0.4} & 74.7\scriptsize{$\pm$0.3} & 84.4\scriptsize{$\pm$0.9} & 77.9\scriptsize{$\pm$1.2} \\ 
FLAC \cite{sarridis2023flac} & 95.4\scriptsize{$\pm$0.3} & 91.6\scriptsize{$\pm$0.5} & 93.2\scriptsize{$\pm$0.3} & 87.2\scriptsize{$\pm$0.7} \\ 
\midrule         
\METHOD & \textbf{96.5\scriptsize{$\pm$0.2}} & \textbf{95.1\scriptsize{$\pm$0.4}} & \textbf{95.3\scriptsize{$\pm$0.5}} & \textbf{92.7\scriptsize{$\pm$1.1}} \\
\bottomrule
\end{tabular}
\end{table}

\section{Ablation Study}
In this section, we explore the ways of integrating bias-capturing features into the training process. Table \ref{tab:abl1} presents a comparison of \METHOD's performance when the bias-capturing features are added to the main features versus when they are concatenated with them. As one may observe, the concatenation approach is much less effective than the addition. This is anticipated, as relying on $\mathbf{b}$, with non-zero corresponding weights, would perform poorly on balanced settings (random background color), while not relying on it, with $\sim0$ corresponding weights, would be equivalent to the sub-optimal vanilla training.
Furthermore, Tab.~\ref{tab:abl2} demonstrates how the selection of layer to incorporate the bias-capturing features affects the performance of \METHOD. The penultimate layer yields the most favorable performance, as the shallower the selected layer, the fewer layers remain independent of the protected attributes. Consequently, the model's capacity to learn fair features gradually diminishes. 

\begin{table}[h]
\begin{center}
\caption{Addition vs Concatenation: Biased-MNIST performance comparison between different approaches of integrating bias-capturing features. }\label{tab:abl1}
    \begin{tabular}{lcccc} 
    \toprule
          \multirow{ 2}{*}{Method}  &    \multicolumn{4}{c}{$q$}     \\ \cline{2-5}
 & 0.99 & 0.995 & 0.997 & 0.999 \\ \midrule
 Concatenation   & 91.5 & 81.7 & 70.3 & 36.5 \\ 
         Addition & \textbf{98.1} & \textbf{97.3} & \textbf{96.3} & \textbf{91.7}\\
\bottomrule

\end{tabular}

\end{center}
\end{table}

\begin{table}[h]
\begin{center}
\caption{\METHOD\ performance on Biased-MNIST with $q=0.99$ when considering different layers for incorporating the bias capturing features. }\label{tab:abl2}
    \begin{tabular}{lcccc} 
    \toprule
          \multirow{ 2}{*}{Method}  &    \multicolumn{4}{c}{Layer}     \\ \cline{2-5}
 & 1st & 2nd & 3rd & 4th \\ \midrule
         \METHOD &  74.6 & 85.8  & 97.7& \textbf{98.1} \\
\bottomrule
\end{tabular}

\end{center}
\end{table}


\section{Conclusion}

In this work, we propose a method for bias mitigation in CV deep-learning models, termed \METHOD. The proposed method essentially enriches the feature representations of a model with bias-capturing features in order to force the model parameters' updates to rely only on unbiased samples, thus leading to fair representations. Through a comprehensive experimental evaluation, we show that the proposed approach surpasses the state-of-the-art in single- as well as multi-attribute bias scenarios. The main limitation of \METHOD\ is the requirement to have access to protected attribute labels, either directly within the dataset or through another dataset where these labels are available.

\section*{Acknowledgments}
This research was supported by the EU Horizon Europe project
MAMMOth (Grant Agreement 101070285).

\bibliographystyle{unsrt}  
\bibliography{templateArxiv}

\end{document}